\documentclass[runningheads,a4paper]{llncs}

\usepackage{amssymb}
\usepackage{graphicx}
\usepackage{subcaption}
\usepackage{multirow}
\usepackage{bbding}
\usepackage{amsmath}
\usepackage{arydshln}
\usepackage{url}
\urldef{\mailsa}\path|anjany.sekuboyina@tum.de|
\urldef{\mailsb}\path||

\begin{document}

\mainmatter  

\title{A Localisation-Segmentation Approach for Multi-label Annotation of Lumbar Vertebrae using Deep Nets}

\titlerunning{Multi-Class Lumbar Vertebrae Segmentation}

%
%
\newcommand*\samethanks[1][\value{footnote}]{\footnotemark[#1]}
\author{Anjany Sekuboyina$^{1,2,}$\Envelope \and Alexander Valentinitsch$^{2}$\and Jan S. Kirschke$^{2}$ \and \\Bjoern H. Menze$^{1}$}
%
\authorrunning{Sekuboyina et al.}

\institute{$^{1}$Technische Universit\"{a}t M\"{u}nchen, Munich, Germany\\ $^{2}$Klinikum rechts der Isar, Munich, Germany \\
\mailsa\\
\mailsb\\}


%
%

\toctitle{Multi-Class Lumbar Vertebrae Segmentation}
\maketitle

\vspace{-0.3in}
\begin{abstract}
Multi-class segmentation of vertebrae is a non-trivial task mainly due to the high correlation in the appearance of adjacent vertebrae. Hence, such a task calls for the consideration of both global and local context. Based on this motivation, we propose a two-staged approach that, given a computed tomography dataset of the spine, segments the five lumbar vertebrae and simultaneously labels them. The first stage employs a multi-layered perceptron performing non-linear regression for locating the lumbar region using the global context. The second stage, comprised of a fully-convolutional deep network, exploits the local context in the localised lumbar region to segment and label the lumbar vertebrae in one go. Aided with practical data augmentation for training, our approach is highly generalisable, capable of successfully segmenting both healthy and abnormal  vertebrae (fractured and scoliotic spines). We consistently achieve an average Dice coefficient of over 90\% on a publicly available dataset of the xVertSeg segmentation challenge of MICCAI`16. This is particularly noteworthy because the xVertSeg dataset is beset with severe deformities in the form of vertebral fractures and scoliosis. 
\end{abstract}

\section{Introduction}
The identification, segmentation, and  quantification of structures visible in medical images is a crucial component in the processing of medical image data. In the context of spinal images, segmentation of spine has an immediate diagnostic importance in clinical decisions around fracture detection and inter-vertebral disc pathology. Segmented spines are also used in the bio-mechanical modelling of the spine for load analysis and fracture prediction. Therefore, an automated approach attempting to segment the spine should posses two key features: (1) Highly generalisable in terms of the fields-of-view (FOV) and scanner calibrations, in addition to variability in the spine's curvature, BMD (bone mineral density) distribution, and micro-architecture and (2) Capable of segmenting images from a clinical population that consists of abnormalities such as vertebral fractures, scoliotic, and kyphotic spines. 

A typical analysis pipeline for spinal images consists of three stages:  spine localisation, vertebrae detection, and spine segmentation. The first two steps of localisation and detection are accomplished by basic routines such as shape matching (using generalised Hough transform \cite{klinder09}) and spine-curve detection (using circle detection in axial slices \cite{forsberg14}). This is followed by a segmentation stage which may tackled using statistical mean shape models or atlases, followed by an optimisation routine that adapts the fitted model to account for local variations \cite{kadoury13,korez14}. Such a pipeline has proven to be highly effective in most of the cases. However, there is a limit to the amount of generalisability such model/shape-based approach can offer in clinical cases. Its limit is determined by the robustness of the chosen model and the amount of relaxation it can withstand during the optimisation routine post fit. It is obvious that such models cannot generalise to a fractured vertebra or a deformed spine. In such cases, learning-based approaches offer respite, provided that the data that the approach can learn from is rich and diverse enough. For example, \cite{glocker12} and \cite{suzani15} solve the problem of vertebra detection in arbitrary FOVs using random forests and multi-layer regressors respectively. Chen et al. \cite{chen15} make use of the omni-present convolutional neural networks to detect vertebrae using an altered cost formulation that takes into account the sequence of the vertebrae. However, there is no \emph{end-to-end} approach that handles every problem in the analysis pipeline (localisation, detection, and segmentation) in one go, that is, takes a 3D spine scan as input and generates an annotated and segmented spine volume.

\begin{figure}[t]
  \centering
    \includegraphics[width=0.9\linewidth]{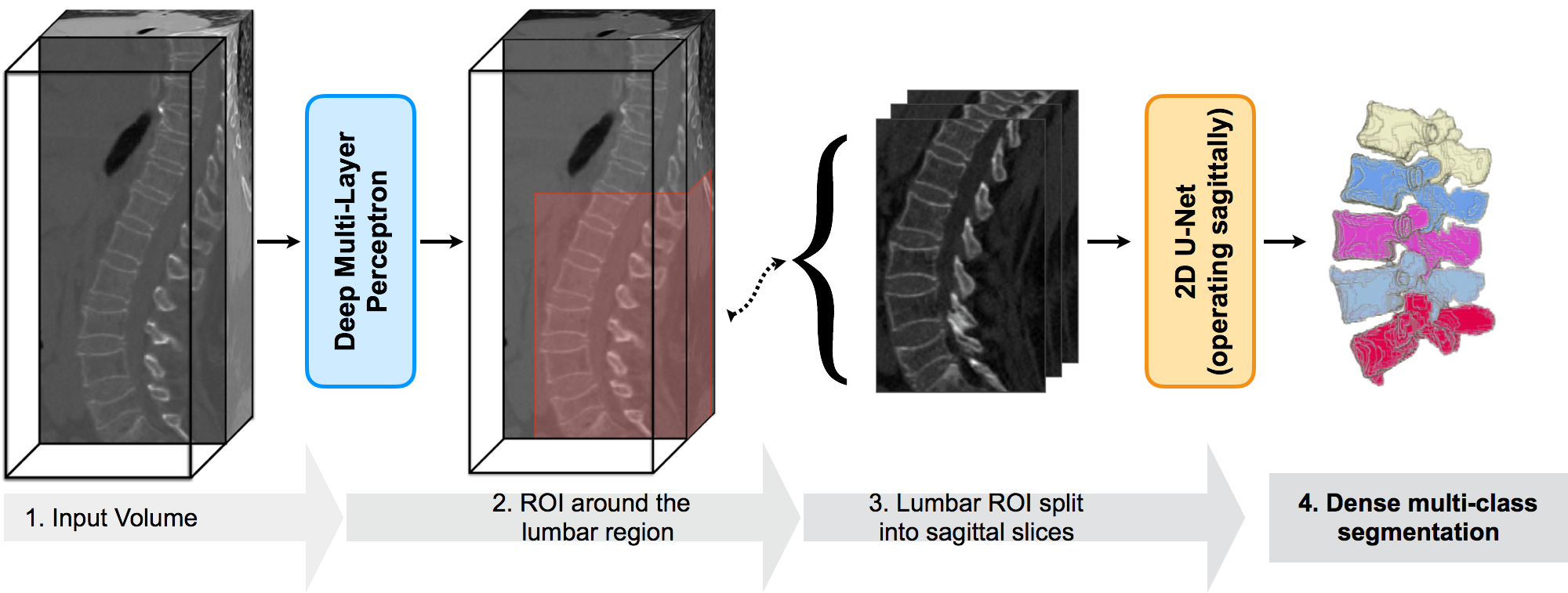} 
  \caption{A schematic diagram giving an overview of our approach.}
   \label{figure:1}
\end{figure}

In this work, we propose an approach that segments and simultaneously labels the the lumbar vertebrae using deep neural networks. Given a CT scan volume of an arbitrary FOV, our approach performs a multi-class segmentation over five classes corresponding to the five lumbar vertebrae (L1 to L5) and a background class. This is done in a two-staged approach: (1) Localise the lumbar region, and (2) Segment the localised lumbar vertebrae in to their respective classes. Both the stages are elaborated in detail in section \ref{sec:methodology}. Figure \ref{figure:1} gives a schematic overview of our approach. We use the dataset released as part of the xVertSeg challenge in MICCAI 2016 to test the performance of our approach. We are the first to attempt this challenge, and achieve a mean Dice coefficient upwards of 90\% on both the training and the test set. Section \ref{sec:exp} contains the implementation and experimental details.

\section{Methodology}
\label{sec:methodology}
\subsection{Lumbar localisation}
\vspace{-2ex}
When compared to typical binary segmentation, multi-class segmentation is inherently difficult due to a more complex representation that is to be learnt. Moreover, the appearance of adjacent vertebrae are highly correlated. Thus, instead of directly attempting the segmentation problem on the entire scan, we choose to restrict our attention to a restricted region of interest - the lumbar region.

\subsubsection{Non-linear Regression using deep neural network} We pose the localisation of the lumbar region as a regression problem, and employ a five-layered perceptron with ReLU (rectified linear unit) activation as a regressor. It is trained on contextual, intensity-dependant features that encode long-range spatial information, as in \cite{glocker12}, and regresses on the location of the six planes that define a bounding box. An $n$ lenght feature, $\mathbf{f}_{i,j,k}$ can be constructed at the voxel location of $(i,j,k)$ as below:
\begin{equation}
\mathbf{f}_{i,j,k} = \{f_{i,j,k}^1, f_{i,j,k}^2, \dots, f_{i,j,k}^n\},
\end{equation}

where $f_{i,j,k}^p$, $\forall p\in{1,2,\dots n}$ is the mean intensity of the 3D image region lying inside a cuboid that is centered at a certain offset from the voxel at $(i,j,k)$. The cuboid's offset and the dimension are generated randomly for construction of a feature. Given these features, each of them corresponding to a voxel location, the regressor should predict the region-of-interest, or a bounding box around the lumbar region.


In a simple set-up, a bounding box can be defined by six planes: $x_{min}$, the smallest $x$-coordinate, $x_{max}$, the largest $x$-coordinate, and their $y$ and $z$ equivalents. These are refereed to as the \emph{bounding planes}. Given the contextual information through the feature $\mathbf{f}_{i,j,k}$, a six-length vector encoding the voxel's location \emph{relative} the bounding planes is learnt, as below:
\begin{equation}
\mathbf{y}_{i,j,k} = \{i-x_{min}, i-x_{max}, j-y_{min}, j-y_{max}, k-z_{min}, k-z_{max}\}
\end{equation}

\subsubsection{Estimation of the lumbar bounding box} 
\begin{figure}[t]
  \centering
  \begin{subfigure}{\linewidth}
    \centering
    \includegraphics[width=0.7\linewidth]{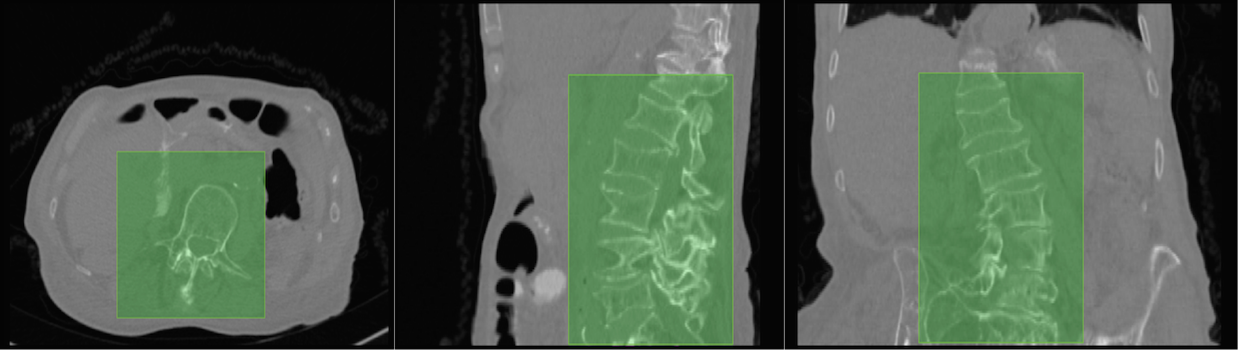}
  \end{subfigure}

  \begin{subfigure}{\linewidth}
    \centering
    \includegraphics[width=0.7\linewidth]{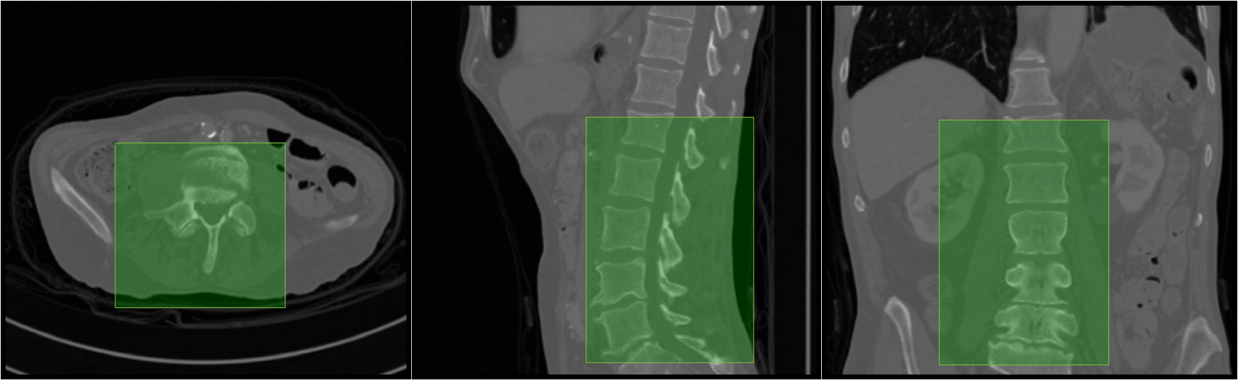}
  \end{subfigure}  
  \caption{Lumbar Localisation: The axial, sagittal, and coronal views of the bounding box localising the lumbar section. (Row 1) Case 18 containing severe abnormalities such as multiple fractures and scoliosis is localised perfectly. (Row 2) Case 22 shows a mild \emph{under}-localisation, not localising a top region of L1.}
   \label{figure:2}
\end{figure}

Each pass through the regressor using a feature corresponding to a certain voxel predicts the locations of the bounding planes with respect to that voxel. In order to speed-up the feature generation procedure without loss of useful information, only the \emph{significant} voxels are considered for feature extraction. For this purpose, the voxels from the response of a Canny's edge detector are used for feature extraction. Thus, every significant voxel votes for a prospective bounding box of which the most representative bounding box is chosen. Figure \ref{figure:2} shows a few examples of the localised lumbar regions.

%
%

\vspace{-3ex}
\subsection{Multi-class Segmentation}
Once the lumbar region is successfully identified, the FOV is restricted, enabling a human to effectively identify the vertebrae, based on certain key points such as the sacrum, number of vertebrae in the FOV etc. We make use of a deep convolutional network to learn such key points on its own in order to segment and annotate the lumbar vertebrae. 
\vspace{-3ex}
\subsubsection{Fully-convolutional network for multi-class segmentation} This is the segmentation of the vertebrae is carried out by a fully convolutional network (FCN). We rely on the 2D U-net \cite{unet2d}, but implement an architecture that one level deeper, i.e., six more convolutional layers, three each in the contracting and expanding path, joined by one additional downsampling and up-convolutional layers, and works on sagittal slices from the localised lumbar region. The motivation for a deeper network is to adapt it towards multi-label classification of vertebrae by increasing the receptive field of the coarsest level. The receptive field of our FCN ($\approx$ 270$\times$270 pixel$^2$ or 27$\times$27$mm^2$) covers at least two vertebrae when working at isotropic $1mm$ resolution. Such a receptive field will force the network to learn the sequence of vertebrae in pairs, L1-L2, L2-L3 etc., so that the sequence of the annotations is always in order.
\vspace{-3ex}
\subsubsection{Pre-training} It is a common practice to pre-train a network for one purpose and use it as an initialisation for another network attempting a related yet more-complex task. For example, Long et al. \cite{long} use the state-of-art recognition networks (VGG-16 etc.) as initialisation for the task of segmentation. On a similar footing, a network trained for binary segmentation of lumbar spine (spine vs. background) is employed as an initialisation for the multi-class segmentation. This alleviates the shortcoming of the limited data at our disposal to train a very deep network for the relatively complex task of multi-class segmentation.
\vspace{-3ex}
\subsubsection{ROI Augmentation} Another key concept in our segmentation routine is the ROI augmentation step. However, the localisations are not \emph{uniform} as shown in figure \ref{figure:2}. There could either be non-lumbar vertebrae showing up in the sagittal slices (usually T11 and T12 in our experiments), or part of the lumbar region could be missing (usually L1 in our experiments). The high correlation in the appearance of the vertebrae makes this problem detrimental. Therefore, in addition to augmenting the sagittal slices using elastic and rigid transformations, we also augment based on varying bounding boxes sizes. Let $h\times w \times d$ be the dimension of the lumbar bounding box obtained from the localisation stage. We augment the sagittal slices from bounding box by randomly choosing a $\delta \in \{5,10,15,20,25\}$ so that the sagittal slice dimensions vary between $h \times w$ and $h-2\delta \times w$. This makes the segmentation network robust to improper lumbar localisations.
\vspace{-3ex}
\subsubsection{Extracting the final segmentations} Once all the sagittal slices are segmented, the final segmented volume is coronally corrected by closing the holes in every label of a coronal slice using morphological closing operation. Finally, a 3D connected-component analysis is performed on each label to discard the smaller connected components. This cleans-up a few stray segmentations in the final volume.  


\section{Experiments and Discussion}
\label{sec:exp}
\vspace{-1ex}
We make use of a publicly available dataset for evaluating the performance of the lumbar localisation and the multi-class segmentation stages of our approach.
\vspace{-5ex}
\subsubsection{xVertSeg Dataset}
The xVertSeg dataset, released as part of the xVertSeg challenge \cite{xvertseg} in MICCAI 2016, consists of fifteen train CT volumes with ground truth segmentations of the lumbar vertebrae (into five classes, L1-L5 ) and ten test CT volumes. The participants do not have access to the ground truth segmentations of the test set. The data is very rich in terms of varying FOVs, spine curvatures, and vertebral deformities.\\
\emph{Ground truth for localisation:} The first and the last slices in the three directions (sagittal, coronal, and axial) consisting of a label were considered to be the bounding planes. A tolerance of 15 slices was added on all sides of the bounding box to prevent a tight cropping of the lumbar region. This expanded bounding box was used as the ground truth for training the localisation network.\\
\emph{Ground truth segmentations for test cases:}  As the challenge organisers did not make the performance metrics of our approach available yet, we opted for an in-house ground truth generation. The near-perfect segmentation from our approach was given to two clinical experts (Rater-1 and Rater-2) for correction. Rater-1 was tasked to correct the entire volume, while Rater-2 was tasked to pick a random subset of sagittal, coronal, and axial slices from a volume and segment them entirely.
\vspace{-3ex}
\subsubsection{Lumbar localisation}
Inspired from \cite{suzani15}, a five-layered neural network is cast as a regressor to map the features ($\mathbf{f}_{i,j,k}$) to the offsets of the bounding planes ($\mathbf{y}_{i,j,k}$). The input layer has $n$ (=500) neurons, followed by four hidden layers with 350, 250, 150, and 50 neurons respectively. The output layer has six neurons corresponding to the offsets of the six bounding planes. All the neurons are ReLUs. The network was implemented in the Caffe \cite{caffe} framework. A squared-error loss was optimised using stochastic gradient descent. The available data was augmented with rigid and elastic transformations. The network was trained for 1000 epochs over a few hours with a learning rate of 1e-3 and a momentum of 0.9. The most representative bounding box is chosen using kernel density estimation, aided by Botev bandwidth \cite{botev} selection. 

To measure the performance of localisation, a measure of \emph{sensitivity} (or true positive rate) was used, as defined:
$
S = 1 - \frac{|\mathcal{G}\cap\mathcal{B}^c|}{\mathcal{|G|}}, 
$
where $\mathcal{G}$ is the of set voxels in the ground truth segmentation, and $\mathcal{B}$ is the set of voxels within the bounding box. We use the ground truth segmentation for Rater-1 for this purpose. The sensitivity measures on the test set are shown in table \ref{table:1}, with a few cases shown visually in figure \ref{figure:2}. We obtain a near perfect localisation of 1.0 in all cases except one (Case025). In order to completely cover the lumbar region, a tolerance of 15 voxels is added to the bounding boxes on all sides before considering the localisation for for the next stage.\\
\emph{The curious case of Case025:} Case025 of the test set is peculiar due to the presence of the entire sacrum (S1, S2, and S3) within the FOV. It is the only such case in the train and the test data. The network thus furnishes an imperfect localisation from L2 to S1. Such a behaviour can be easily rectified with additional representative training data.
\vspace{-2ex}
\begin{table}
\centering
 \begin{tabular}{| c | c | c | c | c | c | c | c | c | c | c |} 
 \hline
 \small{Case16} &  \small{Case17} &  \small{Case18} &  \small{Case19} &  \small{Case20} &  \small{Case21} &  \small{Case22} &  \small{Case23} &  \small{Case24} &  \small{Case25} &  \small{\textbf{Mean}}\\ [0.5ex] 
 \hline
 \hline
 0.98 & 1.0 & 0.99 & 0.99 & 1.0 & 0.99 & 0.98 & 1.0 & 0.98 & 0.94 & \textbf{0.98}\\ [1ex] 
 \hline
\end{tabular}
\vspace{0.1in}
\caption{Sensitivities of the localisation algorithms on the ten test cases. The localisation is near perfect ($\sim$1.0) for all the case except Case25.}
\label{table:1}
\end{table}
\vspace{-10ex}
\subsubsection{Multi-class segmentation}
\begin{figure}[t]
  \centering
    \begin{minipage}[c]{0.5\textwidth}
    \includegraphics[width=1\linewidth]{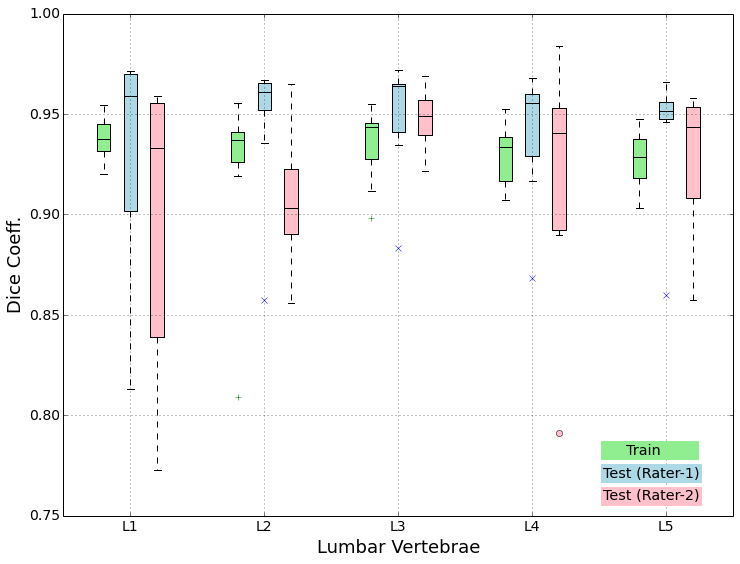} 
    \end{minipage}\hfill
  \begin{minipage}[c]{0.45\textwidth}
  \caption{Plot showing the distribution of the Dice coefficients across the vertebrae, comparing the performance among the datasets.}
   \label{figure:3}
  \end{minipage}
\end{figure}

The segmentation network is implemented in the Caffe framework. A cross-entropy loss is optimised using an Adam solver with an initial learning rate of 1e-4. The binary segmentation network is run for 3000 epochs and the multi-class segmentation (with pre-training) net was run for 2000 epochs. The segmented bounding boxes are reinstated into the actual volumes to obtain the full-resolution segmentations. We report the Dice coefficient for each of the five vertebrae and for the entire lumbar region in table \ref{table:2}. The evaluation is carried out based on the available ground truth segmentations of the train set and those from both Rater-1 and Rater-2 in case of the test set. We also observe a mean Dice score of $\sim$92\%. Since our segmentation is the starting point for Rater-1, a bias in the corresponding performance scores can be observed, with a mean Dice score of 94\%. Figure \ref{figure:3} shows the spread of the Dice coefficients across vertebrae and among the datasets. Observe that the vertebrae in the middle (L3 and L4) are segmented well compared to the peripheral vertebrae (L1 and L5). This is expected since the uncertainty that the net has to overcome for deciding between L1 \& T12 or L5 \& S1 is higher compared to deciding between L2 \& L3 or L3 \& L4 owing to the large receptive field etc.

\vspace{-3ex}
\subsubsection{Discussion}
\begin{figure}[t]
  \centering
    \includegraphics[width=1\linewidth]{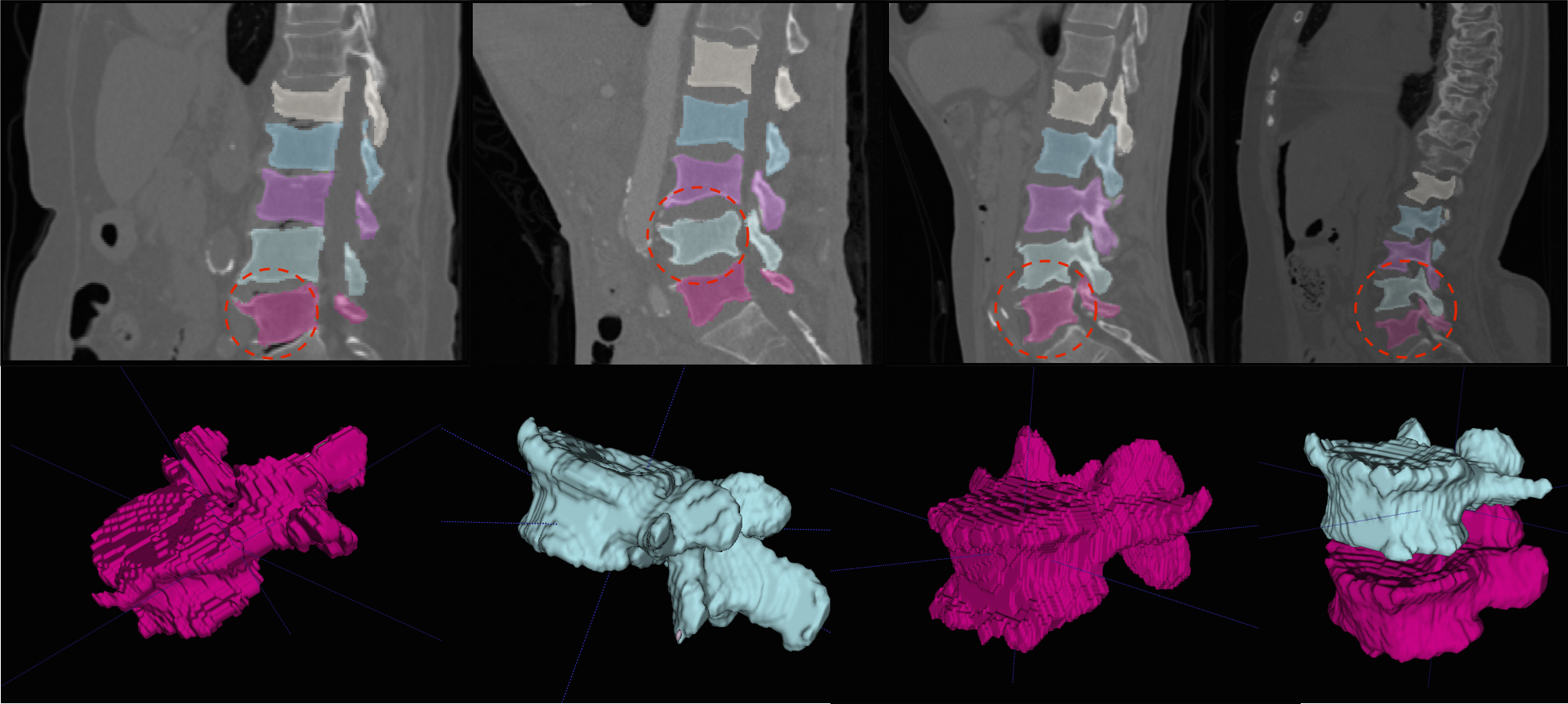} 
  \caption{Multi-class Segmentation: (Row 1) Four sagittal slices where the deformed vertebrae are highlighted, with (Row 2) the 3D rendering of the deformed vertebrae for better visualisation. More visualisations on deformed vertebrae will be made available as supplementary material.}
   \label{figure:4}
\end{figure}

In general, both the stages in our pipeline work remarkably well as per the quantitative results in tables \ref{table:1} and \ref{table:2}. We obtain a near perfect localisation of 1.0 for almost every case, and a mean Dice score of 92\%. In addition to this, the prime motivation of our approach is to successfully segment the deformed spines where the model-based approaches fail. This can be observed visually in figure \ref{figure:4}. Four test cases as shown highlighting the highly deformed spine and vertebrae. Observe that our algorithm successfully segments these cases in spite of the severe deformations.
\vspace{-3ex}
\bgroup
\def\arraystretch{1}
\setlength\tabcolsep{0.05in}
\begin{table}
\centering
 \begin{tabular}{| c | c | c | c | c | c | c |} 
 \hline
 \rule{0pt}{3ex} \textbf{Data} &  \textbf{L1} &  \textbf{L2} &  \textbf{L3} & \textbf{L4} & \textbf{L5} & \textbf{Lumbar} \\ [0.5ex] 
 \hline
 \hline
 \rule{0pt}{3ex}Train & 92.6$\pm$6.7  & 92.7$\pm$3.0  & 93.5$\pm$1.5  & 92.9$\pm$1.3  & 92.7$\pm$1.2  & \textbf{92.7}$\pm$\textbf{2.3}  \\ [1ex]
 \hdashline
 \rule{0pt}{3ex}Test (Rater-1) & 92.7$\pm$5.5  & 94.9$\pm$3.2  & 95.1$\pm$2.5  & 94.2$\pm$2.9  & 94.1$\pm$2.9  & \textbf{94.3}$\pm$\textbf{2.8} \\ [1ex] 
 Test (Rater-2) & 89.7$\pm$6.7  & 90.8$\pm$3.2  & 94.8$\pm$1.5  & 91.8$\pm$5.6  & 92.8$\pm$3.3  & \textbf{92.0}$\pm$\textbf{2.3} \\ [1ex] 
 \hline
\end{tabular}
\vspace{0.1in}
\caption{Dice coefficients (in \%) of our approach on the xVertSeg dataset. Observe a consistent performance of above 90\% in Dice scores. The distribution of the scores is visualised in figure \ref{figure:3}}
\label{table:2}
\end{table}
\egroup
%
%

\vspace{-6ex}
\section{Conclusions}
\label{sec:conc}
Deep-learning based algorithms are a way forward if generalisability is to be achieved. However, usage of such algorithms for dense segmentation is still in its incipient stage. The task of segmentation becomes more challenging when it involves multiple-classes over similar-looking vertebrae. We propose a two-staged approach with deep networks for localisation of the lumbar region and segmenting it into multiple classes. We are the first to present results on the xVertSeg dataset, with an incredible performance achieving a mean Dice coefficient of above 90\%. We also highlight the ability of our approach to handle severe deformities in the spine, which prior approaches would struggle with. We believe that our approach can form a basis for handling more complicated tasks of multi-class segmentation of the entire spine.



\section{Appendix}

\begin{figure*}[!t]
\centering
   \begin{subfigure}[b]{0.75\textwidth}
   \includegraphics[width=1\linewidth]{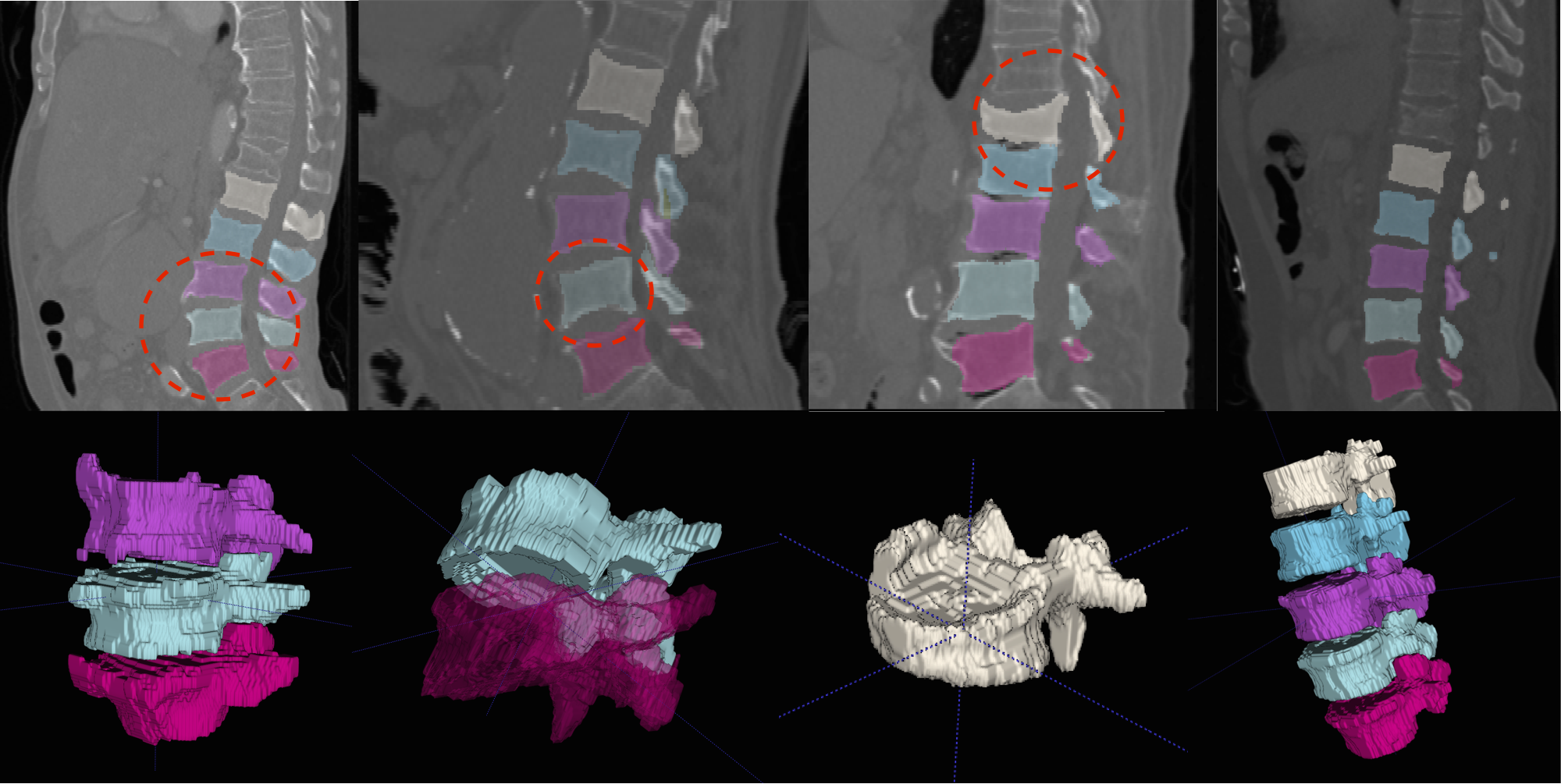}
   \caption{}
   \label{fig:Ng1} 
\end{subfigure}

\begin{subfigure}[b]{0.6\textwidth}
   \includegraphics[width=1\linewidth]{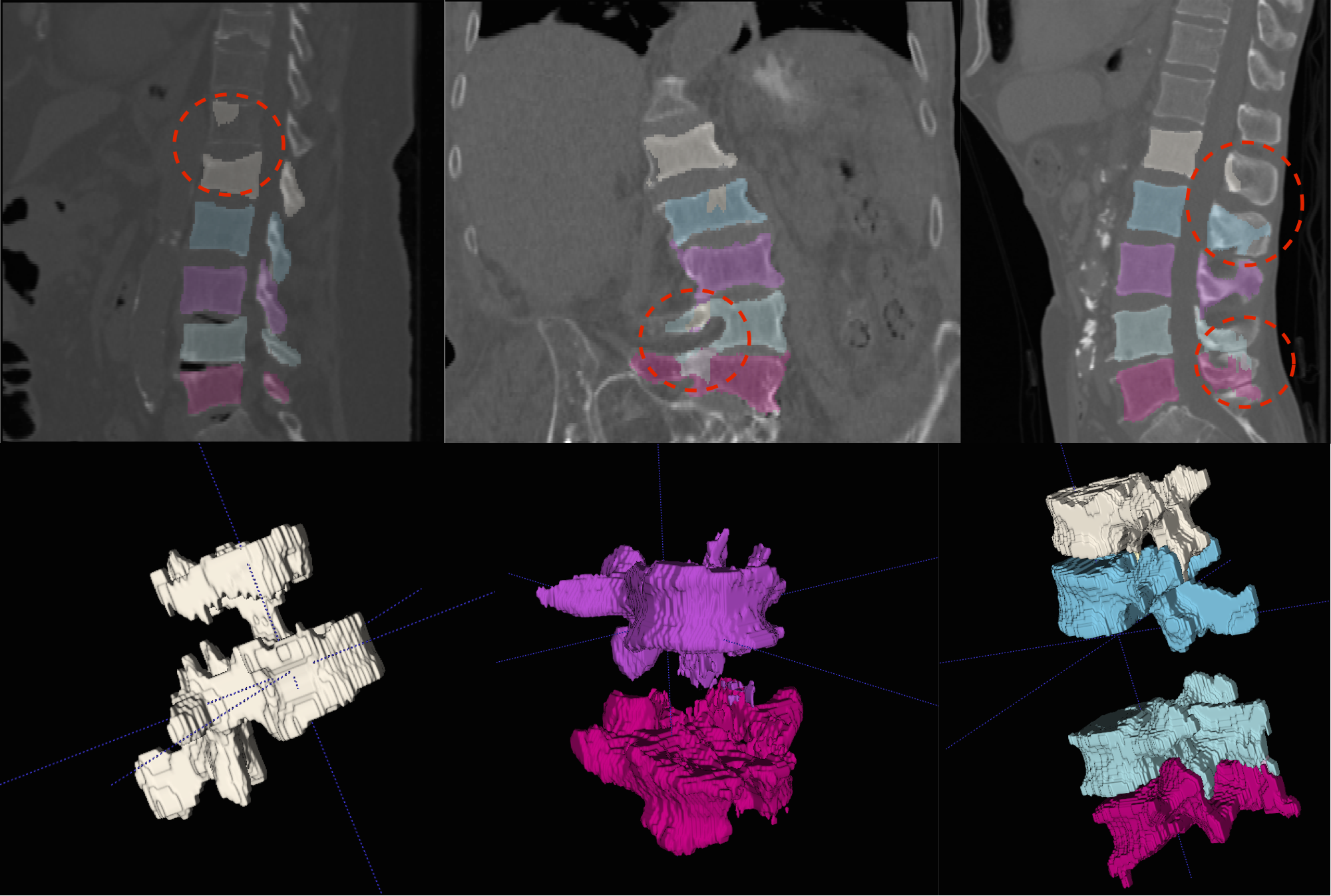}
   \caption{}
   \label{fig:Ng2}
\end{subfigure}
\caption{{\textbf{(a)} Segmented lumbar region (Row 1) with a 3D rendering of the highlighted deformities (Row 2) such as osteophytes and fractures that are successfully segmented. \textbf{(b)} Slightly aberrant cases: (Left) The fracture in L4-L5 is perfectly segmented. However, notice an over-segmentation in L1. (Centre) A successful segmentation of a severely scoliotic spine and deformed vertebrae. Notice some non-homogeneity in segmented labels. Also observe in 3D, a well-captured crush in L3 and an unsegmented region in L5. (Right) The anterior regions of the vertebrae are successfully segmented. However, posterior regions of L1, L2, L4, and L5 are not fully segmented.}}
\end{figure*}

\subsection{Additional Results} 
We present more results of multi-class segmentation on the test set of xVertSeg (figure 5) in addition to the results in figure 4, thereby emphasising the robustness of our approach. We also present a few aberrant segmentations analysing which could further improve our approach.

\subsection{The Outlier (Case 25 of the xVertSeg Dataset)  } As mentioned in section 3 (Lumbar localisation) of the main article, the localisation in Case 25 occurs with a sensitivity of 0.94 (figure 6(c), red outline) as it is the only example in the train and test data that consists of three sacral bones (S1, S2, and S3) within the field-of-view (figure 6(a)). When the scan, with S3 manually cropped off, is used as input (figure 6(b)), the lumbar localisation is perfect (sensitivity of 1.0), as shown by the green outline in figure 6(c). It is clear that the improper localisation is a consequence of working with limited data, and can easily be averted by increasing the variability in the training dataset.
\begin{figure*}
\centering
   \includegraphics[width=0.8\linewidth]{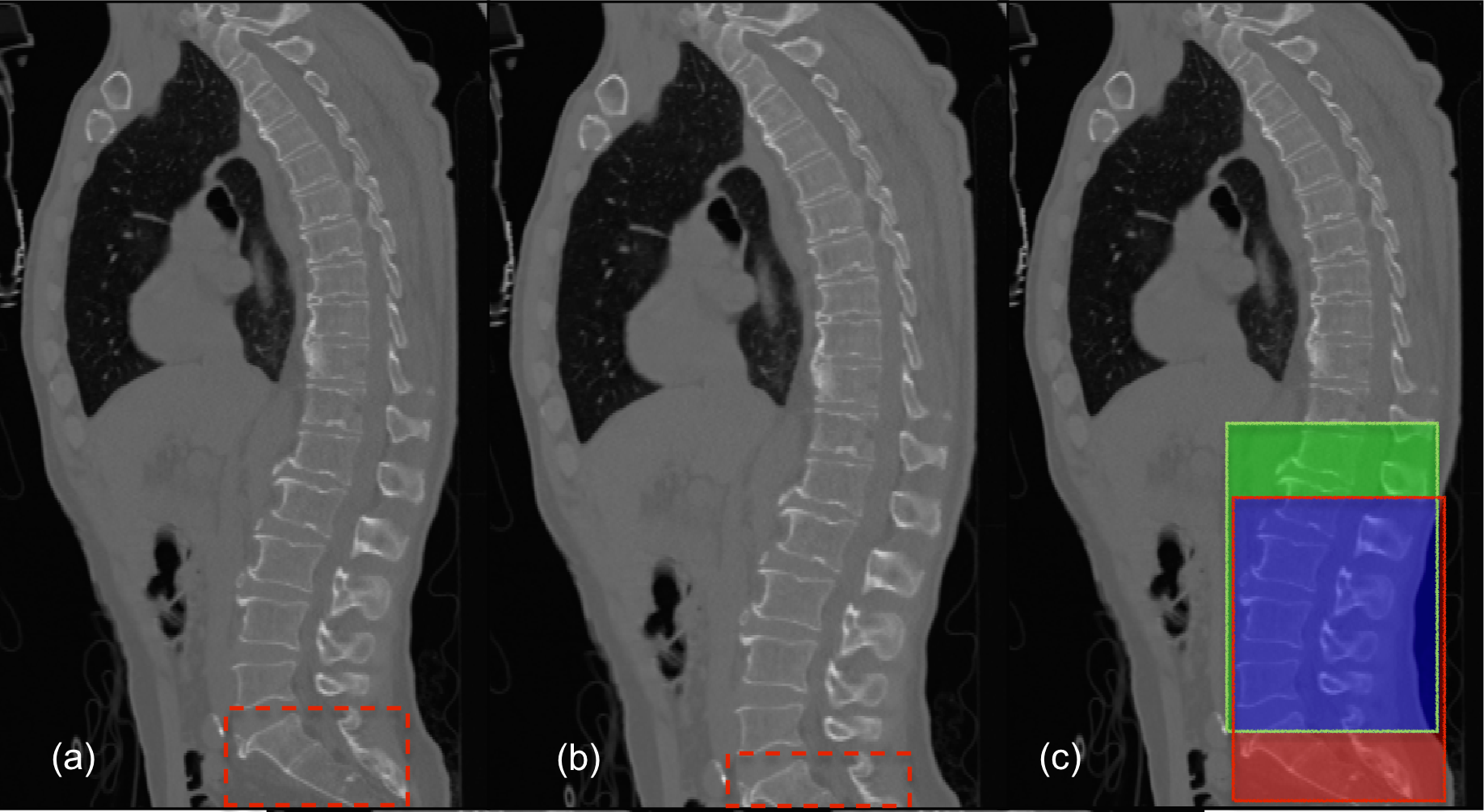}
\caption{{\textbf{(a)} Actual Case 25. \textbf{(b)} Case 25 with S3 cropped off. \textbf{(c)} Red and green regions correspond to lumbar localisation using (a) and (b) respectively. Blue is the overlap in the regions. Both the bounding boxes are overlaid on the original image.}} 
\end{figure*}

\end{document}